# Convolutional Support Vector Machine

Wei-Chang Yeh
Integration and Collaboration Laboratory
Department of Industrial Engineering and Engineering Management
National Tsing Hua University
yeh@ieee.org

The support vector machine (SVM) and deep learning (e.g., convolutional neural networks (CNNs)) are the two most famous algorithms in small and big data, respectively. Nonetheless, smaller datasets may be very important, costly, and not easy to obtain in a short time. This paper proposes a novel convolutional SVM (CSVM) that has both advantages of CNN and SVM to improve the accuracy and effectiveness of mining smaller datasets. The proposed CSVM adapts the convolution product from CNN to learn new information hidden deeply in the datasets. In addition, it uses a modified simplified swarm optimization (SSO) to help train the CSVM to update classifiers, and then the traditional SVM is implemented as the fitness for the SSO to estimate the accuracy. To evaluate the performance of the proposed CSVM, experiments were conducted to test five well-known benchmark databases for the classification problem. Numerical experiments compared favorably with those obtained using SVM, 3-layer artificial NN (ANN), and 4-layer ANN. The results of these experiments verify that the proposed CSVM with the proposed SSO can effectively increase classification accuracy.

## 1. INTRODUCTION

Data mining is an effective method for examining and learning from extensive compound datasets of varying quality [1], and has been broadly applied to numerous practical problems in medicine [2, 3, 4], engineering [5], time series data [6], image classification [7], speech recognition [8], handwritten recognition [9], management [10], and social sciences [11], with classification being one of the most popular topics in data mining. Numerous classifiers for data mining have been established such as support vector machines (SVMs) [3, 4, 7] and deep learning algorithms [8, 9, 12].

Deep learning based on artificial neural networks (ANNs) is made up of neurons that have learnable weights and biases such that the neural network, a special mathematical function, is connected or close to the data in dataset as much as possible [12, 13]. Deep learning techniques include convolution neural networks (CNNs) for the continuous space data types (e.g., image and speech recognition) [7, 8, 14]; recurrent neural networks (RNNs) for the time series data types (e.g., stock markets and language modeling) [12]; generative adversarial networks (GANs) for generating new examples and classifying examples [15]. Deep learning is an adequate and straightforward data-mining method for big data [12, 13]. Moreover, since deep learning techniques need big data to learn the classification rules, that is, they only work well for large datasets, they pose an enormous challenge to many applications with respect to obtaining large enough datasets [12, 13]. Furthermore, deep learning relies on good hardware, especially the graphics processing unit (GPU), to have better performance, but such hardware is still expensive [16, 17].

The SVM is another well-known and effective supervised learning model for selecting attributes and classifying data. Before the rise of deep learning, the SVM outperformed ANNs in various real-life applications such as in the medicine [3, 4], semiconductor industry [18], on-line analysis [19], spectral unmixing resolution [20], imbalanced datasets [21], etc [22, 23, 24]. In comparison with deep learning techniques that try to connect data in terms of ANNs, the SVM separates (not to connect) different classes of data based on the kernels through mathematical optimization [25, 26]. In addition, an SVM has high accuracy with less computation power and small data, which are two shortcomings of deep learning [22, 23, 24]. Therefore, besides the original SVM, various enhanced SVMs have been developed before the development deep learning [21, 22, 23, 24]. SVMs are discussed in detail in Section 5.1.

Small data are well-formatted data with small volumes that are accessible, understandable, and actionable for decision makers [27]. The value of data lies in the information content, but not the volume of data [28]. For some cases such as the marketing strategies of targeting campaigns or delivering personalized experiences, big data might not be appropriate because they do not require full-on big data [29]. Conversely, small data extract an individual's data and provide valuable information to help decision makers formulate strategies. Moreover, the occurrence of small data is rare, with the process of collecting them being expensive and strenuous [4, 21]. Hence, if the data mining of small data is improved, it will aid in making useful, cost-efficient, and timely decisions in small data applications.

Deep learning techniques and SVMs belong to a broader family of machine learning algorithms. Deep learning techniques (e.g., convolution neural networks (CNNs)) based on neural networks are powerful for mining big data, but less effective in smaller datasets. On the contrary, SVMs outperform all neural network types in smaller datasets, but are less effective in mining big data. This paper proposes a novel convolutional SVM (CSVM) that has the advantages of both SVM and deep learning to enhance SVM by maximizing its prediction accuracy, and tests for classifying two-class datasets.

The proposed CSVM employs a supervised learning technique that is based on simplified swarm optimization (SSO), which is another powerful machine learning algorithm [2, 6, 30, 31, 32, 33]. Numerical experiments and comparative research with ANNs and the traditional SVM show the accuracy and effectiveness of the proposed CSVM tested on five two-class datasets.

## 2. PROPOSED AND TRADITIONAL CONVOLUTION PRODUCTS

The major difference between the proposed CSVM and the traditional SVM is the convolution product. Hence, the traditional and the proposed new convolution products are introduced and discussed in Section 2.

### 2.1 Convolution-related concept

CNNs represent some of the most significant models of deep learning, and their performance has been verified in numerous recognition research areas. Among the vital operation techniques of CNNs, we introduce some that are used in this paper [12, 13].

1. Padding: To prevent the reduction in data size generated by the convolution process in the next layer, we add zeros around the input image, with such action being called padding.
2. Stride: A kernel that is moving a horizontal or vertical distance each time is called a stride. The greater the stride is, the more independent the neighboring values in the convolution process.
3. Convolution: In each operation of convolution, multiplication of the values between the input and the

kernel (filter) moves through based on the given stride after padding. Then, these products are summed up and filled in the corresponding positions on the next layer.

**2.2 Proposed convolution product with repeated attributes**

Suppose that $N_{att}$, $N_{sol}$, $N_{filter}$, and $N_{var}$ are the numbers of attributions, solutions, filters constructed in each solution, and the variables contained in each filter, respectively. Let $v_{r,a} = v_{r,a,0}$ be the value of the $a$th attribute in the $r$th record and $v_{r,a,f}$ be the value of the $a$th attribute in the $r$th record after using the $f$th filter, where $a = 1, 2, \ldots, N_{att}$, $s = 1, 2, \ldots, N_{sol}$, and $f = 1, 2, \ldots, N_{filter}$. For example, nine attributes are used in the breast cancer dataset of University of California Irvine (UCI) [34], and the vector representing the first normalized record is listed below:

$$I_1 = [0.50, 0.10, 0.10, 0.10, 0.20, 0.10, 0.30, 0.10, 0.10]$$
$$= [v_{1,1}, v_{1,2}, v_{1,3}, v_{1,4}, v_{1,5}, v_{1,6}, v_{1,7}, v_{1,8}, v_{1,9}]. \quad (1)$$

The $v_{r,a,f}$ is calculated using the convolution product in terms of $v_{r,a,f-1}, v_{r,a+1,f-1}, \ldots, v_{r,a+N_{var}-1,f-1}$, and the filter $X_{s,f}$ as follows:

$$v_{r,a,f} = v_{r,a,f-1} \times x_{s,f,1} + v_{r,a+1,f-1} \times x_{s,f,2} + \ldots + v_{r,a+N_{var}-1,f-1} \times x_{s,f,N_{var}}. \quad (2)$$

From Eq. (2), there are $N_{var}$ attributes that are included in the $a$th attribute if $l < N_{att}$. However, no attributes $v_{r,N_{att}+1}, v_{r,N_{att}+2}, \ldots, v_{r,N_{att}+N_{var}-1}$ are included when we need to use Eq. (2) to update the last $l$ attributes with $l < N_{var}$.

Let filter $X_{s,1} = [x_{s,1,1}, x_{s,1,2}, x_{s,1,3}] = [-1, 0, 1]$. The procedures for generating the new attributes using the convolution product are listed as follows.

$$v_{1,1,1} = v_{1,1} \times x_{s,1,1} + v_{1,2} \times x_{s,1,2} + v_{1,3} \times x_{s,1,3} = 0.50 \times (-1) + 0.10 \times 0 + 0.10 \times 1 = -0.4 \quad (3)$$

$$v_{1,2,1} = v_{1,2} \times x_{s,1,1} + v_{1,3} \times x_{s,1,2} + v_{1,4} \times x_{s,1,3} = 0.10 \times (-1) + 0.10 \times 0 + 0.10 \times 1 = 0.0 \quad (4)$$

$$v_{1,3,1} = v_{1,3} \times x_{s,1,1} + v_{1,4} \times x_{s,1,2} + v_{1,5} \times x_{s,1,3} = 0.10 \times (-1) + 0.10 \times 0 + 0.20 \times 1 = 0.1, \quad (5)$$

$$v_{1,4,1} = v_{1,4} \times x_{s,1,1} + v_{1,5} \times x_{s,1,2} + v_{1,6} \times x_{s,1,3} = 0.10 \times (-1) + 0.10 \times 0 + 0.20 \times 1 = 0.0, \quad (6)$$

$$v_{1,5,1} = v_{1,5} \times x_{s,1,1} + v_{1,6} \times x_{s,1,2} + v_{1,7} \times x_{s,1,3} = 0.10 \times (-1) + 0.10 \times 0 + 0.20 \times 1 = 0.1, \tag{7}$$

$$v_{1,6,1} = v_{1,6} \times x_{s,1,1} + v_{1,7} \times x_{s,1,2} + v_{1,8} \times x_{s,1,3} = 0.10 \times (-1) + 0.10 \times 0 + 0.20 \times 1 = 0.0, \tag{8}$$

$$v_{1,7,1} = v_{1,7} \times x_{s,1,1} + v_{1,8} \times x_{s,1,2} + v_{1,9} \times x_{s,1,3} = 0.30 \times (-1) + 0.10 \times 0 + 0.10 \times 1 = -0.10. \tag{9}$$

From the above, the first and second old attributes (i.e., $v_{1,1}$ and $v_{1,2}$) are used only once as shown in Eq. (3) and twice as shown in Eqs. (3) and (4) for generating $v_{1,1,1}$ and $v_{1,2,1}$, respectively. Similarly, the last and the last second attributes in $I_1$, (i.e., $v_{1,9}$ and $v_{1,8}$) are only showed in Eq. (9) and Eqs. (8) and (9), respectively. Moreover, there are no new attributes $v_{1,8,1}$ and $v_{1,9,1}$ based on Eq. (2).

There is no padding in the proposed CSVM. However, we need to guarantee that the following two situations are satisfied to fix the above problems:

1) each attribute is included in the same number (i.e., $N_{var}$) of convolution products,

2) the last $j$ attributes still exist after each convolution product.

The first ($N_{var}-1$) attributes are repeated and appended in the last attribute of the same record such that the total number of attributes is an integer multiple of $N_{var}$, that is, $v_{r,N_{att}+a,f} = v_{r,a,f-1}$ for $a = 1, 2, \ldots, N_{var}-1$ and $f = 1, 2, \ldots, N_{filter}$. Hence, following the same example discussed above, we have

$$v_{1,8,1} = v_{1,8} \times x_{s,1,1} + v_{1,9} \times x_{s,1,2} + v_{1,1} \times x_{s,1,3} = 0.10 \times (-1) + 0.10 \times 0 + 0.50 \times 1 = 0.4, \tag{10}$$

$$v_{1,9,1} = v_{1,9} \times x_{s,1,1} + v_{1,1} \times x_{s,1,2} + v_{1,2} \times x_{s,1,3} = 0.10 \times (-1) + 0.50 \times 0 + 0.10 \times 1 = 0.0. \tag{11}$$

Thus, each new attribute is generated by three convolution products, and we have the new $I_1$ accordingly.

$$I_1 \otimes X_{s,1} = [-0.4, 0.0, 0.1, 0.0, 0.1, 0.0, -0.1, 0.4, 0.0]. \tag{12}$$

Let $I_{i,j}$ be the updated $i$th record after using the $j$th filter, $I_i = I_{i,0}$. The next example demonstrates the updated $I_1$ after two filters are used, with each having $N_{var} = 3$ variables; $X_{s,2} = [x_{s,2,1}, x_{s,2,2}, x_{s,2,3}] = [1.8, -0.9, 0.7]$.

$$v_{1,1,2} = v_{1,1,1} \times x_{s,2,1} + v_{1,2,1} \times x_{s,2,2} + v_{1,3,1} \times x_{s,2,3} = -0.4 \times (1.8) + 0.0 \times (-0.9) + 0.1 \times (0.7) = -0.65, \tag{13}$$

$$v_{1,2,2} = v_{1,2,1} \times x_{s,2,1} + v_{1,3,1} \times x_{s,2,2} + v_{1,4,1} \times x_{s,2,3} = 0.0 \times (1.8) + 0.1 \times (-0.9) + 0.0 \times (0.7) = -0.09, \tag{14}$$

$$v_{1,3,2} = v_{1,3,1} \times x_{s,2,1} + v_{1,4,1} \times x_{s,2,2} + v_{1,5,1} \times x_{s,2,3} = 0.1 \times (1.8) + 0.0 \times (-0.9) + 0.1 \times (0.7) = 0.25, \tag{15}$$

$$v_{1,4,2} = v_{1,4,1} \times x_{s,2,1} + v_{1,5,1} \times x_{s,2,2} + v_{1,6,1} \times x_{s,2,3} = 0.0 \times (1.8) + 0.1 \times (-0.9) + 0.0 \times (0.7) = -0.09, \quad (16)$$

$$v_{1,5,2} = v_{1,5,1} \times x_{s,2,1} + v_{1,6,1} \times x_{s,2,2} + v_{1,7,1} \times x_{s,2,3} = 0.1 \times (1.8) + 0.0 \times (-0.9) - 0.1 \times (0.7) = 0.11, \quad (17)$$

$$v_{1,6,2} = v_{1,6,1} \times x_{s,2,1} + v_{1,7,1} \times x_{s,2,2} + v_{1,8,1} \times x_{s,2,3} = 0.0 \times (1.8) - 0.10 \times (-0.9) + 0.4 \times (0.7) = 0.37, \quad (18)$$

$$v_{1,7,2} = v_{1,7,1} \times x_{s,2,1} + v_{1,8,1} \times x_{s,2,2} + v_{1,9,1} \times x_{s,2,3} = -0.1 \times (1.8) + 0.4 \times (-0.9) + 0.0 \times (0.7) = -0.54, \quad (19)$$

$$v_{1,8,2} = v_{1,8,1} \times x_{s,2,1} + v_{1,9,1} \times x_{s,2,2} + v_{1,1,1} \times x_{s,2,3} = 0.4 \times (1.8) + 0.0 \times (-0.9) - 0.4 \times (0.7) = 0.44, \quad (20)$$

$$v_{1,9,2} = v_{1,9,1} \times x_{s,2,1} + v_{1,1,1} \times x_{s,2,2} + v_{2,1,1} \times x_{s,2,3} = 0.0 \times (1.8) - 0.4 \times (-0.9) + 0.0 \times (0.7) = 0.36. \quad (21)$$

Thus, after using the two filters $X_{s,1}$ and $X_{s,2}$, we have

$$I_r^* = I_1 \otimes X_s = (I_1 \otimes X_{s,1}) \otimes X_{s,2} = [-0.65, -0.09, 0.25, -0.09, 0.11, 0.37, -0.54, 0.44, 0.36] \quad (22)$$

The basic idea of the proposed convolution product with repeated attributes is that the first ($N_{var} - 1$) attributes are repeated and appended in the last attribute of each (updated) record such that the total number of attributes is an integer multiple of $N_{var}$, that is, $v_{r,N_{att}+a,f} = v_{r,a,f-1}$ for $a = 1, 2, \ldots, N_{var}-1$. The pseudo code of the proposed convolution product with repeated attributes is listed as follows:

**Input:** The $r$th record $I_r = [v_{r,1}, v_{r,2}, \ldots, v_{r,N_{att}}]$ and the $s$th solution $X_s$.

**Output:** The $I_r \otimes X_s$.

**STEP C0.** Let $f = 1$ and $v_{r,i,0} = v_{r,i}$ for $i = 1, 2, \ldots, N_{att}$.

**STEP C1.** Let $a = 1$, $v_{r,i,f-1} = v_{r,k,f-1}$ for $i = N_{att} + 1, 2, \ldots, N_{att} + N_{var} - 1$, $k = i - N_{att}$.

**STEP C2.** Let $b = 0$, $i = a$, and $j = 1$.

**STEP C3.** Let $b = b + v_{r,i,f-1} \times x_{s,f,j}$.

**STEP C4.** If $j < N_{var}$, let $i = i + 1$, $j = j + 1$, and go to STEP C3.

**STEP C5.** If $a < N_{att}$, let $v_{r,a,f} = b$, $a = a + 1$, and go to STEP C2.

**STEP C6.** If $f < N_{filter}$, let $f = f + 1$ and go to STEP C1.

Additionally, we obtain the following properties after employing the proposed convolution product with repeated attributes:

**Property 1.** If $x_{s,f,1} = \alpha$ and $x_{s,f,k} = 0$ for all $k = 2, \ldots, N_{var}$ and all $f = 1, \ldots, N_{filter}$, then

$$v_{r,a,f} = \alpha \cdot v_{i,a,f-1} = \alpha^f \cdot v_{i,a,f-1} \qquad (23)$$

for all $a = 2, \ldots, N_{att}$ and $f = 1, \ldots, N_{filter}$.

## 3. PROPOSED AND TRADITIONAL SSO

In the proposed CSVM, all values in filters of the proposed convolution product with repeated attributes are updated based on the proposed new SSO. The traditional SSO is introduced briefly, and the proposed SSO including the new self-adaptive solution structure with *pFilter*, the novel one-solution one-filter one-variable greedy update mechanism, and the fitness function are presented in Section 3.

### 3.1 Traditional SSO

The SSO is one of the simplest machine-learning methods [2, 6, 30, 31, 32, 33] in terms of its update mechanism. It was first proposed by Yeh, and has been tested to be a very useful and efficient algorithm for optimization problems [31, 33], including data mining [2, 6]. Owing to its simplicity and efficiency, SSO is used here to find the best values in filters of the proposed CSVM.

The basic idea of SSO is that each variable, such as the *j*th variable in the *i*th solution $x_{i,j}$, needs to be updated based on the following stepwise function [2, 6, 30, 31, 32, 33]:

$$x_{i,j} = \begin{cases} g_j & \text{if } \rho_{[0,1]} \in [0, C_g) \\ p_{i,j} & \text{if } \rho_{[0,1]} \in [C_g, C_g) \\ x_{i,j} & \text{if } \rho_{[0,1]} \in [C_p, C_w) \\ x & \text{otherwise} \end{cases} \qquad (24)$$

where the value $\rho_{[0,1]} \in [0, 1]$ is generated randomly, the parameters $C_g$, $C_p - C_g$, $C_w - C_p$, $1 - C_w$ are all in [0, 1] and are the probabilities of the current variable that are copied and pasted from the best of all solutions, the best *i*th solution, the current solution, and a random generated feasible value, respectively.

There are different variants of the traditional SSO that are customized to different problems from the no free lunch theorem; for example, the four items in Eq. (24) are also reduced to three items to increase the efficiency; parameters $C_g$, $C_p$, and $C_w$ are all self-adapted; special values or equations are implemented to replace $g_j$, $p_{i,j}$, $x_{i,j}$, and $x$; or only a certain number of variables is selected to be updated, etc. However, the SSO update mechanism is always based on the stepwise function.

### 3.2 Fitness function

Fitness functions help solutions learn toward optimization to attain goals in artificial intelligence, such as the proposed CSVM, the traditional SVM, and the CNN. The accuracy obtained by the SVM, based on the records transferred from the proposed convolutions, is adopted here to represent the fitness to maximize in the CSVM:

$$F(X_i) = \frac{\text{the total number of instances predicted correctly based on } X_i}{\text{the total number of instances}}. \qquad (25)$$

**Input:** All records $I_r = [v_{r,1}, v_{r,2}, \ldots, v_{r,N_{att}}]$ and the $s$th solution $X_s$ for $r = 1, 2, \ldots, N_{rec}$.

**Output:** The $F(X_s)$.

**STEP F0.** Calculate $I_r^* = I_r \otimes X_s$ based on the pseudo code provided in Section 2.2 for $r = 1, 2, \ldots, N_{rec}$.

**STEP F1.** Classifier $\{I_1^*, I_2^*, \ldots, I_{N_{rec}}^*\}$ using the SVM and let the accuracy be $F(X_s)$.

### 3.3 Self-adaptive solution structure and *pFilter*

In the proposed CSVM, each variable of all filters in each solution is initialized randomly from [-2, 2]. Each filter and solution is presented by $N_{var} \times 1$ and $N_{filter} \times N_{var}$, respectively, since the number of filters may be more than one. For example, the $s$th solution $X_s$ and the $f$th filter $X_{s,f}$ in $X_s$ are denoted as follows:

$$X_s = \begin{bmatrix} x_{s,1,1} & x_{s,1,2} & \cdots & x_{s,1,N_{var}} \\ x_{s,2,1} & x_{s,2,2} & \cdots & x_{s,2,N_{var}} \\ \vdots & \vdots & \vdots & \vdots \\ x_{s,N_{filter},1} & x_{s,N_{filter},2} & \cdots & x_{s,N_{filter},N_{var}} \end{bmatrix} = \begin{bmatrix} X_{s,1} \\ X_{s,2} \\ \vdots \\ X_{s,N_{filter}} \end{bmatrix}, \qquad (26)$$

where

$$X_{s,f} = [x_{s,f,1}, x_{s,f,2}, \ldots, x_{s,f,N_{var}}]. \qquad (27)$$

However, overall, the number of filters is equal, that is, $N_{filter}$ for each solution and all generations. However, a greater number of filters does not always guarantee a better fitness value. Hence, we need to record the best number of filters for each solution. Let filter $j$ be the best filter of solution $s = 1, 2, \ldots, N_{sol}$, and define $pFilter[s] = j$ if $F[X_{s,f}] \leq F[X_{s,j}]$ for all $k = 1, 2, \ldots, N_{filter}$. Note that $X_{h,i}$ is the best solution for $pFilter[h]=i$ among all existing solutions if $F[X_{s,f}] \leq F[X_{h,i}]$ for all $s = 1, 2, \ldots, N_{sol}$ and $f = 1, 2, \ldots, N_{filter}$.

In the end, only the best solution (e.g., $X_s$) and its best number of filters, namely, $X_{s,1}, X_{s,2}, \ldots, X_{s,j}$ where $pFilter[s] = j$, are reported. In addition, the update mechanism is based on the best filter in the proposed CSVM. Hence, the solution is self-adapted by the best number of filters.

### 3.4 One-solution one-filter one-variable greedy SSO update mechanism

The proposed new one-solution one-filter one-variable greedy SSO update mechanism is discussed in this subsection.

#### 3.4.1 One-*pFilter* is selected randomly to be updated in each generation

In the proposed CSVM, all values in filters are variables that must be determined to implement convolution products. Without the help from the GPU, it takes a long time to update variables to deepen the SVM. Hence, instead of the traditional algorithms, including SSO, the genetic algorithm (GA), particle swarm optimization (PSO), of which all solutions need to be updated, only one solution is selected randomly for updating in each generation of the proposed new SSO update mechanism. Let solution $s$ be selected to be updated based on the following equations:

$$s = \begin{cases} gBest & \text{if } \rho_{[0,1]} \in [0, C_{g,1}) \\ \rho_{[1,Nsol]} & \text{if } \rho_{[0,1]} \in [C_{g,1}, C_{p,1}) \\ s & \text{if } \rho_{[0,1]} \in [C_{p,1}, C_{w,1}) \\ 0 & \text{otherwise} \end{cases} \tag{28}$$

where $\rho_{[0,1]}$ is a random floating-point number generated from interval $[0, 1]$ and $\rho_{[1,Nsol]} \in \{1, 2, \ldots, N_{sol}\}$ is the index of the solution selected randomly, *gBest* is the index of the best solution found, and the 0 is a new solution generated randomly. The new updated solution $X_s$ will be either discarded or replaced with the old $X_s$ based on the process described next.

### 3.4.2 One-filter one-variable greedy update mechanism

All variables need to be updated, namely, the all-variable update mechanism, in the traditional SSO, and it has a higher probability of escaping the local trap compared to the updates with only some variables. However, the all-variable update mechanism may cause solutions that are near optimums to be kept away from their current positions. Additionally, its runtime is $N_{sol}$ times that of the one-variable update, which selects one variable randomly to be updated. Hence, to reduce the runtime, only one variable in one filter in the solution selected in Section 3.4.1 is updated.

Let $s$ be the solution selected to be updated. In the proposed new SSO, only one filter; for example, $f$ where $f = 1, 2, \ldots, pFilter[f] = j$, in solution $s$ is chosen randomly. Moreover, one variable; for example, $x_{s,f,k}$ where $k = 1, 2, \ldots, N_{var}$, in such filter $X_{s,f}$ is also selected randomly to be updated based on the following simple process:

$$Max\{Min\{x_{s,f,k} + 0.05 \cdot \rho_{[-0.01, 0.04]}, 2\}, -2\}. \tag{29}$$

After resetting all variables in these filters $X_{s,h}$ to a random number generated from [-2, 2] for all $h > f$, we have

$$x^*_{s,h,l} = \begin{cases} x_{s,h,l} & \text{if } (h < f) \text{ or } (h = f \text{ and } l \neq k) \\ Max\{Min\{x_{s,h,l} + 0.05 \cdot \rho_{[-0.01, 0.04]}, 2\}, -2\} & \text{if } h = f \text{ and } l = k \\ \rho_{[-2,2]} & \text{otherwise} \end{cases}. \tag{30}$$

Also, $F[X_{s,l}] = F[X_{s,f-1}]$ for all for all $l < f$.

Moreover, the updated solutions $X_s^*$, including these new updated variables and filters, are all discarded, if their fitness value are not better than that of $X_s$, i.e.,

$$X_s = \begin{cases} X_s & \text{if } F(X_s) > F(X_s^*) \\ X^* & \text{otherwise} \end{cases}. \tag{31}$$

### 3.5 Pseudocode of the proposed SSO

The pseudocode of the proposed SSO based on the new self-adaptive solution structure, *pFilter*, and the new update mechanism are listed as follows:

**Input:** A random selected solution (e.g., $X_s$) with its *pFilter*.

**Output:** The updated $X_s$.

**STEP U0.** Generate a random number $\rho_{[0, 1]}$ from [0, 1] and select a solution, say $X_s$ where $s \in \{1, 2, \ldots, N_{sol}\}$ based on Eq.(28).

**STEP U1.** Select a filter, say $X_{s,j}$ where $j \in \{1, 2, \ldots, pFilter[s]\}$.

**STEP U2.** Update $X_s$ to $X^*$ based on Eq. (30).

**STEP U3.** Based on Eq. (31) to decide to let $X_s = X^*$ or discard $X^*$.

**STEP U4.** If $X_s = X^*$, let $pFilter[s] = f$, where $F(X_i^*) \leq F(X_f^*)$ for all $i = 1, 2, \ldots, N_{filter}$. Otherwise, halt.

**STEP U5.** If $F(X_{gBest, pFilter[gBest]}) \leq F(X_{s, pFilter[s]})$, let $gBest = s$.

## 4. PROPOSED SMALL-SAMPLE OA TO TUNE PARAMETERS

It is important to select the most representative combination of parameters to find good results for all algorithms, such as the three parameters $C_g$, $C_p$, and $C_w$ in SSO. To reduce the computation burden, a novel concept called small-sample orthogonal array (OA) is proposed in terms of OA test to tune parameters in Section 4.

### 4.1 OA

The design of experiment (DOE) adopts an array design that arranges the tests and factors in rows and columns, respectively, such that rows and columns are independent of each other, and there is only one test level in each factor level [35]. The DOE is able to select better parameters from some representative predefined combinations to reduce test numbers [2, 36].

The Taguchi OA test, first developed by Taguchi [35], is a DOE that is implemented to achieve the objective of this study. OA is denoted by $L_n(a^b)$, where $n = a^{(\lceil \log_a(b+1) \rceil)}$, $a$, and $b$ are the numbers of tries, levels of each factor, and factors, respectively. For example, Table 1 represents an OA denoted by $L_9(3^4)$.

Table 1. An experiment of four factors with three levels using an OA

| Try ID | Factor 1 | Factor 2 | Factor 3 | Factor 4 |
| --- | --- | --- | --- | --- |
| 1 | 1 | 1 | 1 | 1 |
| 2 | 1 | 2 | 2 | 2 |
| 3 | 1 | 3 | 3 | 3 |
| 4 | 2 | 1 | 2 | 3 |
| 5 | 2 | 2 | 3 | 1 |
| 6 | 2 | 3 | 1 | 2 |
| 7 | 3 | 1 | 3 | 2 |
| 8 | 3 | 2 | 1 | 3 |
| 9 | 3 | 3 | 2 | 1 |

From Table 1, we can see that the characteristics of the OA are orthogonal as follows.

1. The number of different levels in each column is equal; for example, numbers, 1, 2, and 3 appear three times in each column in Table 1.

2. All ordered pairs of the two factors for the same test also appear exactly once, e.g., (1, 1), (1, 2), (1, 3), (2, 1), (2, 2), (2, 3), (3, 1), (3, 2), and (3,3) in Columns 1 and 2, 1 and 3, 1 and 4, 2 and 3, 2 and 4, and 3 and 4, to ensure that each level is dispersed evenly in the complete combination of each level of factors.

### 4.2 Proposed small-sample OA

There are three major methods for tuning parameters:

1. The try-and-error method: It implements the tests exhaustively by trying all possible cases to find the one with the better results. It is the simplest, but also the most inefficient one.

2. The parameter-adapted method: It selects and tests some set of parameters from the existing parameters, which are already used in some applications. This method may have some issues with respect to identifying the characteristic of new problems.

3. The DOE: It selects the parameters from the experiment design. Compared to the two aforementioned methods, this method is the most efficient and effective one. However, this method faces an efficiency problem in large datasets or needs to be repeated very often.

Hence, to overcome these aforementioned problems, a novel method called the small-sample OA test is proposed to improve the OA method for tuning parameters. To reduce the runtime, the proposed small-sample OA test only samples few data randomly from the dataset and conducts the OA test on the subsets of such small-sample data to find the best parameters that result in the highest accuracy, the shortest runtime, and/or the largest number of solutions with the maximal number of obtained highest accuracy based on the following three rules:

**Rule 1.** The one with the highest accuracy among all others;

**Rule 2.** The one with the shortest runtime, with a big gap between such runtime and others if there is a tie based on Rule 1;

**Rule 3.** The one with the largest number of solutions that have the highest accuracy if there is a tie based on Rule 2.

Then, this selected parameter set is applied to the rest of the unsampled dataset. The example for this proposed test is provided in Section 6.

## 5. PROPOSED CSVM and TRADITIONAL SVM

The proposed CSVM is a convolutional SVM modified by employing a new convolution product, which is updated based on the proposed new SSO. The traditional SVM is introduced briefly, and then the proposed pseudocode of the proposed CSVM is presented.

## 5.1 Traditional SVM

SVMs are excellent machine learning tools for binary classification cases [25, 26]. The purpose of an SVM is to maximize the margin between two support hyperplanes to separate two classes of data. Let $X = \{z_1 = (x_1, y_1), z_2 = (x_2, y_2), \ldots, z_n = (x_n, y_n)\}$ be a two-class dataset for training. For example, in a linear SVM, a hyperplane is a line, and we want to find the best hyperplane $W^T X + b = 0$ to separate these two classes of data in $X$, where $W$ is the weight vector and $b$ is the bias perpendicular to such hyperplane such that $\|W\|$ is as large as possible. The above linear SVM is a constrained optimization model and can be written as follows [25, 26]:

$$\text{Max} \quad \|W\|$$
$$\text{s.t.} \quad y_i(W^T \cdot x_i + b) \geq 1. \tag{32}$$

After applying the Lagrange multiplier method to the constrained optimization model, the SVM problem is a convex quadratic programming problem that can be presented as follows [25]:

$$\text{Min} \quad \|W\|^2 / 2 - \sum_{i=1}^{l} \lambda_i y_i (W^T \cdot x_i + b) + \sum_{i=1}^{l} \lambda_i$$
$$\text{s.t.} \quad \lambda_i \geq 0 \text{ for all } i = 1, 2, \ldots, n. \tag{33}$$

where $\lambda_i$ is the Lagrange multiplier.

For these high-dimensional data, it is very difficult to find a single linear line to separate two different sets. Hence, these data are mapped into a higher dimensional space using a function that is called the kernel in SVM. Then, a hyperplane can be found to separate the mapped data. Here, we list some popular kernel functions [25, 26]:

$$K(z_i, z_j) = (z_i^T z_j + 1)^p \tag{34}$$
$$K(z_i, z_j) = \exp[-\|z_i - z_j\|^2 / (2\sigma^2)] \tag{35}$$
$$K(z_i, z_j) = \tanh(k z_i^T z_j - \delta). \tag{36}$$

For more details of SVM and its development, refer to [25, 26].

## 5.2 Pseudocode of the proposed CSVM

The pseudocode of the proposed CSVM is described below together with the integration of the proposed convolution product discussed in Subsection 2.2, the proposed SSO introduced in Section 3, and the proposed small-sample OA presented in Subsection 4.2.

**PROCEDURE CSVM0**

**Input:** A dataset.

**Output:** The accuracy of the classifier CSVM.

**STEP 0.** Separate the dataset into $k$ folds randomly, and then select one fold (e.g., the $k^*$th fold of the dataset); for the small-sample OA, it has $N$ tries.

**STEP 1.** Implement CSVM0 $(i, k^*)$ using the $i$th parameter setting on the $k^*$th fold of the dataset for $i$ = 1, 2, …, $N$, and then let the parameter setting of the try (e.g., $i^*$) with the highest accuracy among all $N$ tries.

**STEP 2.** Implement CSVM0 $(i^*, j)$ on the $j$th fold of the dataset using the parameter setting of the $i^*$th try for $j$ = 1, 2, …, $k$.

**PROCEDURE CSVM0($\alpha, \beta$)**

**Input:** The parameter setting in the $\alpha$th try of the small-sample OA and the $\beta$th fold of the dataset.

**Output:** The accuracy.

**STEP W0.** Generate solutions $X_s$ randomly, then calculate $F(X_{s,f})$ based on the proposed convolution product and the SVM. Find $pFilter[s]$ and $gBest$ such that $F(X_{gBest,pFilter[gBest]}) \geq F(X_{s,f})$, where $s$ = 1, 2, …, $N_{sol}$ and $f$ =1, 2, …, $N_{filter}$.

**STEP W1.** Let $t = 1$.

**STEP W2.** Update a randomly selected solution based on the pseudocode of the new SSO provided in Subsection 3.5 and the parameter setting in the $\alpha$th try of OA.

**STEP W3.** Increase the value of $t$ by 1, that is, let $t = t + 1$, and then go to STEP W2 if $t < N_{gen}$.

**STEP W4.** Halt, $F(X_{gBest,pFilter[gBest]})$ is the accuracy, and $X_{gBest,pFilter[gBest]}$ is the classifier.

## 6. EXPERIMENTAL RESULTS AND SUMMARY

There are two experiments: Ex1 and Ex2 in this study. Ex1 is based on the proposed small-sample OA concept to find the parameters $C_g$, $C_p$, $C_w$, $N_{gen}$, $N_{filter}$, and $N_{var}$ in the proposed CSVM. Then, these parameters are employed in Ex2 to conduct an extension test to compare these results with those obtained from the DSCM, SVM, 3-layer ANN, and 4-layer ANN, respectively.

### 6.1 Simulation environment

Four algorithms are developed and adapted in this study including the proposed CSVM, SVM, the 3-layer ANN, and the 4-layer ANN. The proposed CSVM is implemented using Dev C++ Version 5.11 C/C++, and the SVM part is integrated by calling the libsvm library [26] with all default setting parameters. The codes of both the 3-layer and 4-layer ANNs are modified using the source code provided in [37], which is coded in Python and run in Anaconda with epochs = 150, batch_size = 10, loss = 'binary_crossentropy', optimizer = 'adam', activation = 'relu' and 12 neurons in the first hidden layer, and activation = 'sigmoid' in the second hidden layer of the 4-layer ANN. The test environment is: Intel(R) Core(TM) i9-9900K CPU @ 3.60 GHz, 32.0 GB memory and Windows 10 64 bits.

To validate the proposed CSVM, the proposed CSVM was compared with the traditional SVM and the 3-layer and 4-layer ANNs on five well-known datasets: "Australian Credit Approval" (A), "breast-cancer" (B), "diabetes" (D), "fourclass" (F), and "Heart Disease" (H) [34] based on a tenfold cross-validation in Ex2. Summary of the five datasets is provided in Table 2.

**Table 2.** Information and characteristics of the five datasets.

| ID | Full Name | Record Number | Attribute Number | Attribute Characteristics |
|---|---|---|---|---|
| A | Australian Credit Approval | 690 | 14 | Integer, Real |
| B | Breast-cancer | 699 | 10 | Integer, Real |
| D | Diabetes | 768 | 8 | Integer, Real |
| F | Fourclass | 862 | 2 | Integer, Real |
| H | Heart Disease | 270 | 13 | Integer, Real |

Let $\Phi$, T, G, f, and N be the highest accuracy levels obtained in the end, the runtime, the earliest generation that obtained $\Phi$, the number of filters generating $\Phi$, the number of solutions that has $\Phi$, respectively. To be easily recognized, the subscripts 25, 50, 75, 100, avg, max, min, and std represent the related values obtained at the end of the 25$^{th}$, 50$^{th}$, 75$^{th}$, and 100$^{th}$ generations, the average, the maximum, minimum, and the standard deviation, respectively.

## 6.2 Ex1: SMALL-SAMPLE OA test

The orthogonal array used in this study is called $L_9(3^4)$ as shown in Table 3. In $L_9(3^4)$, there are nine tries and four factors: $C = (c_g, c_p, c_w, c_r)$, $N_{sol}$, $N_{var}$, and $N_{filter}$; each factor has three levels as shown in Table 4. The higher the level, the larger related values with the exception of $C$, e.g., in level 1, $N_{sol} = 25$ is smaller than that in level 2. The most distinguishable difference amongst all three levels in $C$ of Table 3 is that level 2 has a higher $c_r$ which is to increase the global search ability while level 3 has the lower value of $c_r$ to enhance the local search ability.

**Table 3.** The $L_9(3^4)$.

| Try | $N_{filter}$ | $N_{sol}$ | $N_{var}$ | $(c_g, c_p, c_w, c_r)$ |
|---|---|---|---|---|
| 1 | 1 | 1 | 1 | 1 |
| 2 | 1 | 2 | 2 | 2 |
| 3 | 1 | 3 | 3 | 3 |
| 4 | 2 | 3 | 2 | 1 |
| 5 | 2 | 3 | 3 | 2 |
| 6 | 2 | 1 | 1 | 3 |
| 7 | 3 | 2 | 3 | 1 |
| 8 | 3 | 3 | 1 | 2 |
| 9 | 3 | 1 | 2 | 3 |

**Table 4.** Levels of all-factor test for selecting the parameters.

| Level Code | $N_{filter}$ | $N_{sol}$ | $N_{var}$ | $C = (c_g, c_p, c_w, c_r)$ |
|---|---|---|---|---|
| 1 | 1 | 25 | $\lceil N_{att}/4 \rceil$ | (0.40, 0.30, 0.20, 0.10) |
| 2 | 3 | 50 | $\lceil 2N_{att}/4 \rceil$ | (0.35, 0.25, 0.15, 0.25) |
| 3 | 4 | 75 | $\lceil 3N_{att}/4 \rceil$ | (0.45, 0.30, 0.20, 0.05) |

The results obtained from the proposed CSVM in terms of the proposed small-sample OA test are listed in Table 5, in which each try is run fifteen times, with the larger the $N_{filter}$, $N_{sol}$, $N_{var}$, and/or $N_{gen}$, the longer the runtime. However, it is not necessary to have better fitness values from Table 5. For example, the best fitness value has already been found in $G_{25}$, namely, $F_{25} = F_{50} = F_{75}$, in all datasets except dataset D whose best fitness value is found in $G_{75}$.

Table 5. Levels of all-factor test for selecting the parameters.

| ID | Try | $T_{25}$ | $G_{25}$ | $f_{25}$ | $N_{25}$ | $100F_{25}$ | $T_{50}$ | $G_{50}$ | $f_{50}$ | $N_{50}$ | $100F_{50}$ | $T_{75}$ | $G_{75}$ | $f_{75}$ | $N_{75}$ | $100F_{75}$ |
|---|---|---|---|---|---|---|---|---|---|---|---|---|---|---|---|---|
| A[2] | 1 | 13.09 | 7 | 1 | 12 | 85.00000 | 26.11 | 19 | 1 | 1 | 86.66666 | 39.07 | 19 | 1 | 1 | 86.66666 |
|  | 2 | 27.68 | 9 | 1 | 1 | 88.33334 | 55.33 | 13 | 1 | 1 | 88.33334 | 82.59 | 28 | 1 | 2 | 88.33334 |
|  | 3 | 42.81 | 4 | 1 | 4 | 88.33334 | 86.04 | 11 | 1 | 5 | 88.33334 | 128.39 | 26 | 1 | 6 | 88.33334 |
|  | 4 | 130.54 | 6 | 3 | 13 | 88.33334 | 260.18 | 11 | 3 | 15 | 88.33334 | 389.33 | 11 | 3 | 15 | 88.33334 |
|  | **5** | **43.44** | 6 | 3 | 1 | **90.00000** | 87.39 | 14 | 3 | 1 | 90.00000 | 131.20 | 14 | 3 | 1 | 90.00000 |
|  | 6 | 84.35 | 7 | 3 | 15 | 86.66666 | 168.38 | 13 | 3 | 2 | 88.33334 | 252.73 | 27 | 3 | 3 | 88.33334 |
|  | 7 | 149.28 | 8 | 5 | 4 | **90.00000** | 298.47 | 15 | 5 | 6 | 90.00000 | 447.50 | 24 | 5 | 7 | 90.00000 |
|  | 8 | 215.83 | 9 | 5 | 14 | 88.33334 | 427.65 | 9 | 5 | 14 | 88.33334 | 637.65 | 21 | 5 | 1 | 88.33334 |
|  | 9 | 113.32 | 9 | 5 | 8 | 88.33334 | 225.84 | 18 | 5 | 11 | 88.33334 | 337.70 | 25 | 5 | 15 | 88.33334 |
| B[2] | 1 | 16.03 | 6 | 1 | 13 | 97.05882 | 31.99 | 6 | 1 | 13 | 97.05882 | 47.83 | 26 | 1 | 14 | 97.05882 |
|  | 2 | 33.41 | 5 | 1 | 15 | 97.05882 | 67.26 | 5 | 1 | 15 | 97.05882 | 101.45 | 5 | 1 | 15 | 97.05882 |
|  | 3 | 51.63 | 6 | 1 | 15 | 97.05882 | 103.01 | 6 | 1 | 15 | 97.05882 | 155.17 | 6 | 1 | 15 | 97.05882 |
|  | 4 | 160.82 | 0 | 3 | 15 | 97.05882 | 324.22 | 0 | 3 | 15 | 97.05882 | 485.32 | 0 | 3 | 15 | 97.05882 |
|  | 5 | 53.08 | 2 | 3 | 15 | 97.05882 | 106.13 | 2 | 3 | 15 | 97.05882 | 159.76 | 2 | 3 | 15 | 97.05882 |
|  | 6 | 103.51 | 0 | 3 | 15 | 97.05882 | 209.98 | 0 | 3 | 15 | 97.05882 | 316.64 | 0 | 3 | 15 | 97.05882 |
|  | 7 | 185.49 | 0 | 5 | 15 | 97.05882 | 372.27 | 17 | 5 | 1 | 98.52941 | 558.38 | 17 | 5 | 1 | 98.52941 |
|  | 8 | 271.51 | 9 | 5 | 1 | **98.52941** | 542.93 | 9 | 5 | 1 | 98.52941 | 815.96 | 9 | 5 | 1 | 98.52941 |
|  | **9** | **91.88** | 5 | 5 | 1 | **98.52941** | 184.50 | 17 | 5 | 2 | 98.52941 | 276.67 | 17 | 5 | 2 | 98.52941 |
| D[1] | 1 | 22.70 | 0 | 1 | 15 | 78.08219 | 45.41 | 0 | 1 | 15 | 78.08219 | 68.09 | 0 | 1 | 15 | 78.08219 |
|  | 2 | 45.36 | 8 | 1 | 8 | 80.82191 | 90.93 | 19 | 1 | 11 | 80.82191 | 136.65 | 21 | 1 | 12 | 80.82191 |
|  | 3 | 68.45 | 4 | 1 | 2 | 82.19178 | 137.29 | 13 | 1 | 2 | 82.19178 | 206.65 | 27 | 1 | 4 | 82.19178 |
|  | 4 | 207.15 | 8 | 3 | 2 | 82.19178 | 416.77 | 19 | 3 | 5 | 82.19178 | 624.94 | 29 | 3 | 6 | 82.19178 |
|  | 5 | 68.42 | 9 | 3 | 3 | 82.19178 | 137.33 | 17 | 3 | 5 | 82.19178 | 206.86 | 29 | 3 | 6 | 82.19178 |
|  | 6 | 136.61 | 9 | 3 | 7 | 80.82191 | 274.81 | 15 | 3 | 10 | 80.82191 | 412.70 | 23 | 3 | 12 | 80.82191 |
|  | **7** | 232.55 | 6 | 5 | 6 | 82.19178 | 464.95 | 18 | 4 | 10 | 82.19178 | 696.72 | 25 | 5 | 1 | **83.56165** |
|  | 8 | 345.07 | 4 | 5 | 15 | 80.82191 | 691.15 | 4 | 5 | 15 | 80.82191 | 1037.06 | 4 | 5 | 15 | 80.82191 |
|  | 9 | 114.58 | 9 | 5 | 1 | 82.19178 | 230.88 | 18 | 5 | 2 | 82.19178 | 347.32 | 29 | 5 | 5 | 82.19178 |
| F[3] | 1 | 19.82 | 0 | 1 | 15 | 80.23256 | 39.71 | 0 | 1 | 15 | 80.23256 | 59.61 | 0 | 1 | 15 | 80.23256 |
|  | 2 | 39.97 | 0 | 1 | 15 | 80.23256 | 80.54 | 0 | 1 | 15 | 80.23256 | 121.10 | 0 | 1 | 15 | 80.23256 |
|  | 3 | 61.63 | 9 | 1 | 12 | **83.72093** | 123.97 | 19 | 1 | 14 | 83.72093 | 186.52 | 22 | 1 | 15 | 83.72093 |
|  | 4 | 181.25 | 0 | 1 | 15 | 80.23256 | 364.80 | 0 | 1 | 15 | 80.23256 | 548.34 | 0 | 1 | 15 | 80.23256 |
|  | **5** | 62.05 | 6 | 3 | **15** | **83.72093** | 124.67 | 6 | 3 | 15 | 83.72093 | 187.85 | 6 | 3 | 15 | 83.72093 |
|  | 6 | 120.07 | 0 | 1 | 15 | 80.23256 | 242.21 | 0 | 1 | 15 | 80.23256 | 364.72 | 0 | 1 | 15 | 80.23256 |
|  | 7 | 211.59 | 2 | 5 | 15 | **83.72093** | 424.27 | 2 | 5 | 15 | 83.72093 | 637.24 | 2 | 5 | 15 | 83.72093 |
|  | 8 | 303.38 | 0 | 1 | 15 | 80.23256 | 610.26 | 0 | 1 | 15 | 80.23256 | 918.33 | 0 | 1 | 15 | 80.23256 |
|  | 9 | 100.15 | 0 | 1 | 15 | 80.23256 | 202.20 | 0 | 1 | 15 | 80.23256 | 304.68 | 0 | 1 | 15 | 80.23256 |
| H[2] | **1** | **12.58** | 8 | 1 | 10 | **88.00000** | 26.11 | 19 | 1 | 11 | 88.00000 | 39.67 | 24 | 1 | 13 | 88.00000 |
|  | 2 | 26.09 | 9 | 1 | 10 | **88.00000** | 54.12 | 19 | 1 | 14 | 88.00000 | 81.41 | 26 | 1 | 15 | 88.00000 |
|  | 3 | 40.98 | 9 | 1 | 5 | **88.00000** | 82.38 | 16 | 1 | 7 | 88.00000 | 124.00 | 28 | 1 | 9 | 88.00000 |
|  | 4 | 122.50 | 6 | 3 | 14 | **88.00000** | 244.67 | 6 | 3 | 14 | 88.00000 | 366.86 | 6 | 3 | 14 | 88.00000 |
|  | 5 | 39.94 | 8 | 3 | 10 | **88.00000** | 80.13 | 8 | 3 | 10 | 88.00000 | 122.23 | 24 | 3 | 12 | 88.00000 |
|  | 6 | 79.05 | 5 | 3 | 12 | **88.00000** | 160.39 | 16 | 3 | 15 | 88.00000 | 241.89 | 16 | 3 | 15 | 88.00000 |
|  | 7 | 136.33 | 8 | 5 | 12 | **88.00000** | 273.14 | 15 | 5 | 13 | 88.00000 | 409.73 | 21 | 5 | 14 | 88.00000 |
|  | 8 | 203.64 | 7 | 5 | 15 | **88.00000** | 407.86 | 7 | 5 | 15 | 88.00000 | 612.67 | 7 | 5 | 15 | 88.00000 |
|  | 9 | 68.08 | 9 | 5 | 8 | **88.00000** | 135.69 | 17 | 5 | 9 | 88.00000 | 203.97 | 25 | 5 | 12 | 88.00000 |

[1]. Select the setting based on Rule 1.
[2]. Select the setting based on Rule 2.
[3]. Select the setting based on Rule 3.

Adhering to Rule 1 listed in Section 4, only the try with the highest accuracy Φ is selected to be used for the rest of the unsampled dataset. In this case, Try 7 is selected for Dataset D, since the greatest accuracy is obtained from Try 7 in $G_{75}$. From Rule 2, the runtime T must be considered if there are two tries tied in accuracy. For example, both Try 5 and Try 9 have the highest accuracy in Dataset A, but Try 5 is selected, since its runtime is only 43.43, which is considerably less than the runtime (149.28) of Try 9. Similarly, Try 7 and Try 1 are selected for Datasets B and H, respectively. The parameter setting for the rest of datasets, namely, Dataset F, is based on Rule 3, and Try 5 is selected in accordance with Rule 2.

Hence, we obtain the parameter settings listed in Table 6.

**Table 6.** Parameters used in the proposed CSVM.

| ID | Try ID | $N_{rec}$ | $N_{att}$ | $N_{filter}$ | $N_{sol}$ | $N_{var}$ | $(c_g, c_p, c_w, c_r)$ | $N_{gen}$ | $100F_{SVM}$ | $100F_{N_{gen}}$ | T |
|---|---|---|---|---|---|---|---|---|---|---|---|
| A | 5 | 690 | 14 | 3 | 75 | $\lceil 3N_{att}/4 \rceil = 11$ | (0.35, 0.25, 0.15, 0.25) | 25 | 81.666664 | 90.00000 | 43.44 |
| B | 9 | 699 | 10 | 4 | 25 | $\lceil 2N_{att}/4 \rceil = 5$ | (0.45, 0.30, 0.20, 0.05) | 25 | 95.588234 | 98.52941 | 91.88 |
| D | 7 | 768 | 8 | 4 | 50 | $\lceil 3N_{att}/4 \rceil = 6$ | (0.40, 0.30, 0.20, 0.10) | 75 | 76.712326 | 83.56165 | 696.72 |
| F | 5 | 862 | 2 | 3 | 75 | $\lceil 3N_{att}/4 \rceil = 2$ | (0.35, 0.25, 0.15, 0.25) | 25 | 80.232559 | 83.72093 | 62.05 |
| H | 1 | 270 | 13 | 1 | 25 | $\lceil N_{att}/4 \rceil = 4$ | (0.40, 0.30, 0.20, 0.10) | 25 | 80.000000 | 88.00000 | 12.58 |

In dataset F, there is only two attributes resulting in also two variables in each filter of Table 6. Another observation is that the values $N_{filter}$, $N_{var}$, $N_{sol}$, and $N_{gen}$ are always the smallest, since all the best final fitness values are equal to 88.00000 regardless of the generation number. Then, the parameter setting with less runtime is selected, which is reasonable. This is similar to dataset B whose solution number is only 25, with less local search ability.

In Table 5, the accuracy levels obtained from SVM for the first fold of each dataset are listed in the last second column named $100F_{SVM}$. From Table 5, all values in $F_{N_{gen}}$ are better than those in the corresponding $F_{SVM}$. Moreover, also from Table 5, all fitness values obtained from $G_{25}$, namely, $F_{25}$, are already at least equal to $F_{SVM}$, that is, $F_{SVM} \leq F_{25} \leq F_{50} \leq F_{75} \leq F_{100}$. Hence, the proposed CSVM outperforms the traditional SVM in the small-sample OA, and the wide discrepancy between the final performances of the CSVM and the SVM is further reinforced in Subsection 6.3 using the parameters setting from the proposed small-sample OA.

## 6.3 Ex2

The results obtained from the proposed CSVM for $G_{25}$, $G_{50}$, $G_{75}$, and $G_{100}$ and from the 3-layer and 4-layer ANNs are marked 1-6, respectively in Fig. 1. The results for $G_{100}$ are collected to evaluate the effectiveness of the concept of the proposed small-sample OA and verify any possible effects on the average and the best fitness values of higher generation numbers. The complete data including the average, best, worst, and standard deviation of fitness of each fold for each dataset are listed in Appendix A.

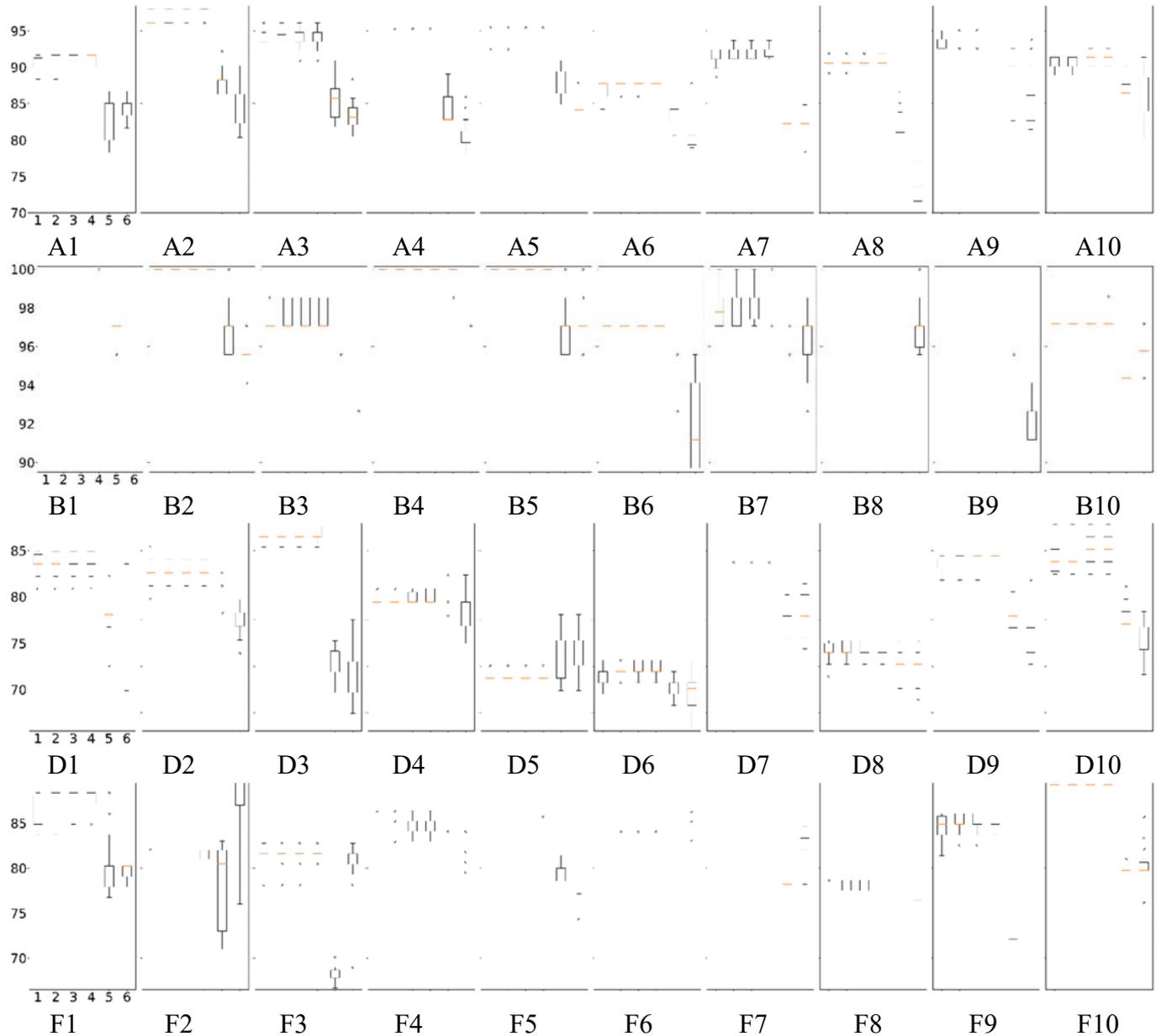

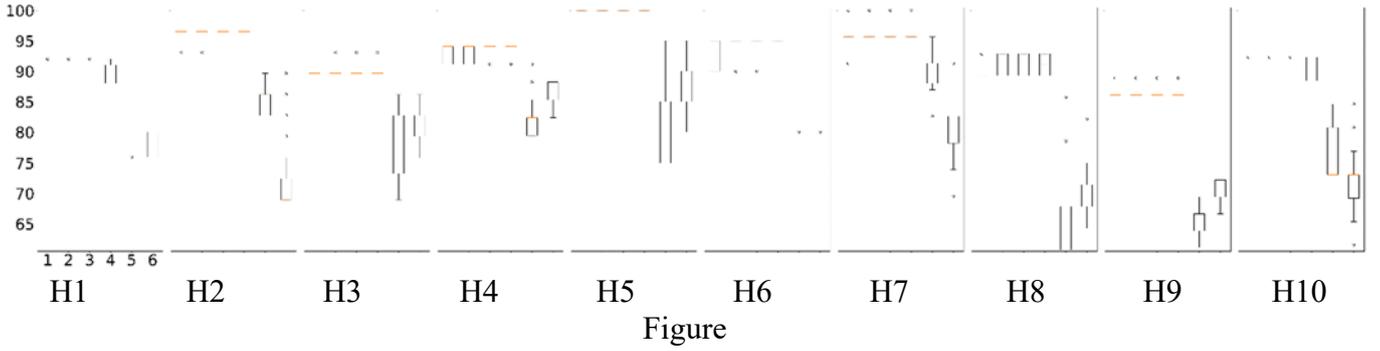

H1　　H2　　H3　　H4　　H5　　H6　　H7　　H8　　H9　　H10

Figure

### 6.3.1 Boxplots of the experimental results from Ex2

From Fig. 1, both results obtained from the 3-layer and 4-layer ANNs are the least favorable with a big gap between the proposed CSVM and the traditional SVM. Hence, these two ANN-based methods are not discussed further, and we only focus on the proposed CSVM and the traditional SVM.

We determined from Fig 1 that the higher the generation number, the better average fitness value. However, it can be observed that the best fitness value remains unchanged from $G_{75}$ to $G_{100}$ except for the 8$^{th}$ fold in Dataset A, the 1$^{st}$ and 10$^{th}$ folds in Dataset B, and the 5$^{th}$ fold in Dataset F. Therefore, $N_{gen} = 75$ is acceptable and there is no need for $N_{gen} = 100$ to increase the fitness value of the best solution, as shown in Fig 1. The position (the fitness values obtained) and the length (the range of the fitness values) of box in $G_{100}$ are frequently higher and shorter than those of $G_{25}$ in most boxplots. Hence, a larger generation number has a higher probability of enhancing the average solution quality under the cost of the longer runtime, but ultimately does little to improve the best fitness value.

### 6.3.2 Number of folds for finding the final best fitness values

Table 7 lists the number of folds that have found the final best fitness values. The subscripts of dataset ID in the first column of Table 7 indicate the generation number used in Ex 2; for example, $B_{25}$ indicates that 25 generations are used for dataset B based on the parameters obtained with small-sample OA in Ex 1. Folds 7, 8, 10, 8, and 10 (see bold numbers in Table 7) under $G_{25}$, $G_{25}$, $G_{75}$, $G_{25}$, and $G_{25}$ in Dataset A, B, D, F, and H, respectively, have found the best final fitness values.

The folds written as subscripts indicate the final best fitness values that have failed to be found. For example, the $7_{1,4,7}$ in ($G_{25}$, $A_{25}$) represents that there are seven folds (from the ten folds) that have already found the best fitness values after 25 generations with the remaining three folds 1, 4, and 7 failing to do so in Dataset A.

Table 7. Number of folds that have found the best solution.

| Dataset | $G_{25}$ | $G_{50}$ | $G_{75}$ | $G_{100}$ |
|---------|----------|----------|----------|-----------|
| $A_{25}$ | **$7_{1,4,7}$** | $8_{8,10}$ | $9_8$ | 10 |
| $B_{25}$ | **$8_{1,10}$** | $8_{1,10}$ | $8_{1,10}$ | 10 |
| $D_{75}$ | $9_7$ | 10 | **10** | 10 |
| $F_{25}$ | **$8_{5,6}$** | $9_5$ | $9_5$ | 10 |
| $H_{25}$ | **10** | 10 | 10 | 10 |

To calculate the probability of the best final fitness value in Table 7, we add the folds (7 + 8 + 10 + 8 +10) and divide the product by the total number of folds (50) in table 6 to get 86%, which informs us that the probability of finding the best final fitness value without reaching $G_{100}$, which entails a significantly longer runtime is 86%.

Hence, the proposed small-sample OA is effective in setting parameters to increase the efficiency and solution quality of the proposed CSVM. The above observation further confirms that having better parameters ultimately negates the need for a greater generation number to increase the fitness of the best solution.

### 6.3.3 ANOVA of the Experimental Results

To investigate the small-sample OA, the Analysis of variance (ANOVA) is carried out to test the average fitness obtained from the proposed CSVM in terms of the parameters set by the small-sample OA, as shown in Table 8. The cells marked with "v" indicate that there is a significant gap between the pair of distinctive generation numbers listed in their respective rows in the fold denoted by the column. This is reinforced through the distinct difference between the average fitness values obtained from $G_{25}$ and $G_{75}$ in all folds of dataset A.

From Table 8, the minimal generation numbers should be 75 and 50 for only Datasets A and F, respectively, with an insignificant gap between the fitness values in each fold. Hence, the proposed small-small OA is still effective in determining the generation number to reduce the significant difference among all fitness values, even it focuses only on the best fitness value and not the average fitness value that we found.

Table 8. The significant differences between two parameter settings.

| | | 1 | 2 | 3 | 4 | 5 | 6 | 7 | 8 | 9 | 10 |
|---|---|---|---|---|---|---|---|---|---|---|---|
| A | (25, 50) | | | | | | | | | | |
| | (25, 75) | v | v | v | v | v | v | v | v | v | v |
| | (25, 100) | v | v | v | v | v | v | v | v | v | v |
| | (50, 75) | | | | | | | | | | |
| | (50, 100) | v | v | v | v | v | v | v | v | v | v |
| | (75, 100) | | | | | | | | | | |
| B | (25, 50) | | | | | | | | | | |
| | (25, 75) | | | | | | | | | | |
| | (25, 100) | | | | | | | | | | |
| | (50, 75) | | | | | | | | | | |
| | (50, 100) | | | | | | | | | | |
| | (75, 100) | | | | | | | | | | |
| D | (25, 50) | | | | | | | | | | |
| | (25, 75) | | | | | | | | | | |
| | (25, 100) | | | | | | | | | | |
| | (50, 75) | | | | | | | | | | |
| | (50, 100) | | | | | | | | | | |
| | (75, 100) | | | | | | | | | | |
| F | (25, 50) | v | v | v | v | v | v | v | v | v | v |
| | (25, 75) | v | v | v | v | v | v | v | v | v | v |
| | (25, 100) | v | v | v | v | v | v | v | v | v | v |
| | (50, 75) | | | | | | | | | | |
| | (50, 100) | | | | | | | | | | |
| | (75, 100) | | | | | | | | | | |
| G | (25, 50) | | | | | | | | | | |
| | (25, 75) | | | | | | | | | | |
| | (25, 100) | | | | | | | | | | |
| | (50, 75) | | | | | | | | | | |
| | (50, 100) | | | | | | | | | | |
| | (75, 100) | | | | | | | | | | |

### 6.3.4 MPI of the Experimental Results

To further investigate the development of the proposed CSVM, two other indices: the average maximum possible improvement ($MPI_{avg}\%$) and the best maximum possible improvement ($MPI_{avg}\%$) are introduced and defined as:

$$MPI_{avg}\% = \text{(the average fitness obtained} - F_{SVM})/ (1 - F_{SVM}) \qquad (37)$$

$$MPI_{max}\% = \text{(the best fitness obtained} - F_{SVM})/ (1 - F_{SVM}). \qquad (38)$$

The $MPI_{avg}\%$ and $MPI_{max}\%$ results are listed in Tables 9 and 10, respectively, where the cells marked "*" indicate that both the related $F_{svm}$ and the average and/or the best fitness obtained are 100% correct; for example, the 2nd, 4th, and 5th folds in dataset B in Table 9. The bold numbers denote the best values among all folds for each dataset under the same generation number. Note that a value of 100, as in the 7th fold of Dataset B in Table 10, indicate that the related accuracy is 100%.

TABLE 9. The 100×MPI$_{avg}$%.

|   |        | 1     | 2     | 3     | 4     | 5     | 6     | 7     | 8     | 9     | 10    | avg   |
|---|--------|-------|-------|-------|-------|-------|-------|-------|-------|-------|-------|-------|
| A | $F_{25}$  | 46.67 | 58.33 | **58.46** | 20.00 | 48.75 | 33.03 | 50.95 | 35.46 | 50.91 | 35.00 | 43.76 |
|   | $F_{50}$  | 49.39 | **65.00** | 61.28 | 20.67 | 51.25 | 35.76 | 54.05 | 37.58 | 53.94 | 36.67 | 46.56 |
|   | $F_{75}$  | 50.61 | **68.33** | 64.36 | 20.67 | 52.50 | 36.06 | 54.76 | 38.18 | 55.76 | 38.06 | 47.93 |
|   | $F_{100}$ | 52.42 | **70.00** | 65.13 | 22.67 | 52.92 | 36.36 | 55.71 | 40.30 | 56.97 | 39.44 | 49.19 |
| B | $F_{25}$  | 66.67 | *     | 38.89 | *     | *     | 50.00 | 52.22 | 0.00  | **75.00** | 0.00  | 40.40 |
|   | $F_{50}$  | 66.67 | *     | 43.33 | *     | *     | 50.00 | 55.56 | 0.00  | **75.00** | 0.00  | 41.51 |
|   | $F_{75}$  | 66.67 | *     | 45.56 | *     | *     | 50.00 | 61.11 | 0.00  | **75.00** | 0.00  | 42.62 |
|   | $F_{100}$ | 67.78 | *     | 48.89 | *     | *     | 50.00 | 62.22 | 0.00  | **75.00** | 1.67  | 43.65 |
| D | $F_{25}$  | **29.22** | 24.58 | 24.37 | 13.12 | 0.16  | 8.97  | 0.00  | 9.09  | 14.00 | 26.04 | 14.96 |
|   | $F_{50}$  | **30.98** | 26.46 | 25.83 | 13.75 | 0.32  | 11.41 | 0.22  | 10.15 | 16.22 | 28.75 | 16.41 |
|   | $F_{75}$  | **31.37** | 27.29 | 27.08 | 14.17 | 0.48  | 12.18 | 0.22  | 11.21 | 17.11 | 30.21 | 17.13 |
|   | $F_{100}$ | **31.37** | 27.71 | 27.08 | 14.37 | 0.48  | 12.18 | 0.44  | 11.82 | 17.56 | 30.62 | 17.36 |
| F | $F_{25}$  | 30.39 | 10.16 | 21.59 | 11.25 | 8.33  | 5.88  | 0.00  | 13.91 | **36.83** | 18.18 | 15.65 |
|   | $F_{50}$  | 34.31 | 11.11 | 23.02 | 12.92 | 8.33  | 6.08  | 0.00  | 14.64 | **39.21** | 18.18 | 16.78 |
|   | $F_{75}$  | 35.49 | 11.59 | 23.33 | 13.96 | 8.33  | 6.27  | 0.00  | 15.07 | **40.00** | 18.18 | 17.22 |
|   | $F_{100}$ | 36.27 | 12.06 | 23.65 | 15.63 | 8.61  | 6.47  | 0.00  | 15.36 | **40.48** | 18.18 | 17.67 |
| H | $F_{25}$  | 42.00 | **78.67** | 40.00 | 38.33 | *     | 67.50 | 75.00 | 5.56  | 44.81 | 26.67 | 46.50 |
|   | $F_{50}$  | 42.67 | **79.33** | 40.67 | 42.50 | *     | 73.33 | 75.83 | 11.11 | 45.56 | 28.33 | 48.81 |
|   | $F_{75}$  | 44.67 | **80.00** | 40.67 | 44.17 | *     | 74.17 | 75.83 | 13.33 | 45.93 | 30.00 | 49.86 |
|   | $F_{100}$ | 45.33 | **80.00** | 40.67 | 44.17 | *     | 75.00 | 75.83 | 16.67 | 45.93 | 32.50 | 50.68 |

TABLE 10. The 100×MPI$_{max}$%.

|   |        | 1     | 2     | 3     | 4     | 5     | 6     | 7     | 8     | 9     | 10    | avg   |
|---|--------|-------|-------|-------|-------|-------|-------|-------|-------|-------|-------|-------|
| A | $F_{25}$  | 54.55 | 75.00 | **76.92** | 20.00 | 62.50 | 36.36 | 57.14 | 45.45 | 63.64 | 41.67 | 53.32 |
|   | $F_{50}$  | 63.64 | 75.00 | **76.92** | 40.00 | 62.50 | 36.36 | 64.29 | 45.45 | 63.64 | 41.67 | 56.95 |
|   | $F_{75}$  | 63.64 | 75.00 | **76.92** | 40.00 | 62.50 | 36.36 | 64.29 | 45.45 | 63.64 | 50.00 | 57.78 |
|   | $F_{100}$ | 63.64 | 75.00 | **76.92** | 40.00 | 62.50 | 36.36 | 64.29 | 54.55 | 63.64 | 50.00 | 58.69 |
| B | $F_{25}$  | 66.67 | *     | 66.67 | *     | *     | 50.00 | **100.00** | 0.00  | 75.00 | 0.00  | 51.19 |
|   | $F_{50}$  | 66.67 | *     | 66.67 | *     | *     | 50.00 | **100.00** | 0.00  | 75.00 | 0.00  | 51.19 |
|   | $F_{75}$  | 66.67 | *     | 66.67 | *     | *     | 50.00 | **100.00** | 0.00  | 75.00 | 0.00  | 51.19 |
|   | $F_{100}$ | **100.00** | * | 66.67 | *   | *     | 50.00 | **100.00** | 0.00  | 75.00 | 50.00 | 63.10 |
| D | $F_{25}$  | 35.29 | 37.50 | 31.25 | 18.75 | 4.76  | 15.38 | 0.00  | 13.64 | 20.00 | **43.75** | 22.03 |
|   | $F_{50}$  | 35.29 | 37.50 | 31.25 | 18.75 | 4.76  | 15.38 | 6.67  | 13.64 | 20.00 | **43.75** | 22.70 |
|   | $F_{75}$  | 35.29 | 37.50 | 31.25 | 18.75 | 4.76  | 15.38 | 6.67  | 13.64 | 20.00 | **43.75** | 22.70 |
|   | $F_{100}$ | 35.29 | 37.50 | 31.25 | 18.75 | 4.76  | 15.38 | 6.67  | 13.64 | 20.00 | **43.75** | 22.70 |
| F | $F_{25}$  | 41.18 | 14.29 | 28.57 | 25.00 | 8.33  | 5.88  | 0.00  | 17.39 | **42.86** | 18.18 | 20.17 |
|   | $F_{50}$  | 41.18 | 14.29 | 28.57 | 25.00 | 8.33  | 11.76 | 0.00  | 17.39 | **42.86** | 18.18 | 20.76 |
|   | $F_{75}$  | 41.18 | 14.29 | 28.57 | 25.00 | 8.33  | 11.76 | 0.00  | 17.39 | **42.86** | 18.18 | 20.76 |
|   | $F_{100}$ | 41.18 | 14.29 | 28.57 | 25.00 | 16.67 | 11.76 | 0.00  | 17.39 | **42.86** | 18.18 | 21.59 |
| H | $F_{25}$  | 60.00 | 80.00 | 40.00 | 50.00 | *     | 75.00 | **100.00** | 33.33 | 55.56 | 50.00 | 60.43 |
|   | $F_{50}$  | 60.00 | 80.00 | 60.00 | 50.00 | *     | 75.00 | **100.00** | 33.33 | 55.56 | 50.00 | 62.65 |
|   | $F_{75}$  | 60.00 | 80.00 | 60.00 | 50.00 | *     | 75.00 | **100.00** | 33.33 | 55.56 | 50.00 | 62.65 |
|   | $F_{100}$ | 60.00 | 80.00 | 60.00 | 50.00 | *     | 75.00 | **100.00** | 33.33 | 55.56 | 50.00 | 62.65 |

As shown in Tables 9 and 10, the results obtained from the proposed CSVM are at least 14.96% and 20.17%, with at most a 50.68% and 63.10% improvement in $MPI_{avg}\%$ and $MPI_{max}\%$, respectively. The results shed light on the effectiveness of the proposed CSVM in comparison with the traditional SVM. It can be also observed that the more attributes, the greater the results obtained from the proposed CSVM regardless of the number of records. Ultimately, compared to the traditional SVM, the proposed CSVM is more suitable for small data.

## 7. CONCLUSIONS AND FUTURE WORK

Classification is of utmost importance in data mining. The proposed new classifier, CSVM, is a convolutional SVM modified with a new repeated-attribute convolution product, in which all variables in each filter are updated and trained based on the proposed novel SSO. Equipped with a self-adaptive structure and *pFilter*, this greedy SSO is a one-solution, one-filter, one-variable type and its parameters are delineated by the proposed small-sample OA.

According to the experiment results for the five UCI datasets, namely, Australian Credit Approval, Breast-cancer, Diabetes, Fourclass, and Heart Disease [34], from Ex2 in Section 6, the proposed CSVM with the parameter setting selected from Ex1 outperforms the traditional SVM, the 3-layer ANN, and the 4-layer ANN with an improved accuracy of at least 14.96% and up to 50.68% in $MPI_{avg}\%$. Hence, the proposed small-sample OA discussed in Subsection 4.2 enables the CSVM to improve its overall performance, while the proposed CSVM ultimately serves as a successful concoction of the advantages of SVM, the convolution product, and SSO.

The classifier design method is a crucial element in the provision of useful information in the modern world. Through comparisons of the results of experiments, it can be determined that whether further research will be conducted on the proposed CSVM, which will be applied to multi-class datasets with more attributes, classes, and records, and amalgamated with particular feature selections.

ACKNOWLEDGMENTS

This research was supported in part by the National Science Council of Taiwan, R.O.C. under grant MOST 104-2221-E-007-061-MY3 and MOST 107-2221-E-007-072-MY3.

# APPENDEX A.

Table A1. The results obtained based on the Australian credit approval dataset.

| Fold | Index | SVM | $F_{25}$ | $F_{50}$ | $F_{75}$ | $F_{100}$ |
|---|---|---|---|---|---|---|
| 1 | AVG | 81.666664 | 90.222222 | 90.722221 | 90.944444 | 91.277776 |
|  | MAX | 81.666664 | 91.666664 | 93.333336 | 93.333336 | 93.333336 |
|  | MIN | 81.666664 | 88.333336 | 88.333336 | 90.000000 | 90.000000 |
|  | STDEV | 0.000000 | 1.048015 | 1.043436 | 1.043437 | 0.947201 |
| 2 | AVG | 92.156860 | 96.732025 | 97.254901 | 97.516339 | 97.647058 |
|  | MAX | 92.156860 | 98.039215 | 98.039215 | 98.039215 | 98.039215 |
|  | MIN | 92.156860 | 96.078430 | 96.078430 | 96.078430 | 96.078430 |
|  | STDEV | 0.000000 | 0.940124 | 0.977006 | 0.881915 | 0.797722 |
| 3 | AVG | 83.116882 | 92.987014 | 93.463204 | 93.982685 | 94.112555 |
|  | MAX | 83.116882 | 96.103897 | 96.103897 | 96.103897 | 96.103897 |
|  | MIN | 83.116882 | 90.909088 | 90.909088 | 90.909088 | 90.909088 |
|  | STDEV | 0.000000 | 1.430810 | 1.426060 | 1.543568 | 1.552959 |
| 4 | AVG | 92.187500 | 93.750000 | 93.802083 | 93.802083 | 93.958333 |
|  | MAX | 92.187500 | 93.750000 | 95.312500 | 95.312500 | 95.312500 |
|  | MIN | 92.187500 | 93.750000 | 93.750000 | 93.750000 | 93.750000 |
|  | STDEV | 0.000000 | 0.000000 | 0.285272 | 0.285272 | 0.540228 |
| 5 | AVG | 87.878792 | 93.787877 | 94.090907 | 94.242422 | 94.292928 |
|  | MAX | 87.878792 | 95.454544 | 95.454544 | 95.454544 | 95.454544 |
|  | MIN | 87.878792 | 92.424240 | 92.424240 | 93.939392 | 93.939392 |
|  | STDEV | 0.000000 | 0.728274 | 0.728274 | 0.616422 | 0.651793 |
| 6 | AVG | 80.701752 | 87.076024 | 87.602340 | 87.660820 | 87.719299 |
|  | MAX | 80.701752 | 87.719299 | 87.719299 | 87.719299 | 87.719299 |
|  | MIN | 80.701752 | 84.210526 | 85.964912 | 85.964912 | 87.719299 |
|  | STDEV | 0.000000 | 0.975533 | 0.445102 | 0.320306 | 0.000000 |
| 7 | AVG | 82.278481 | 91.308018 | 91.856540 | 91.983121 | 92.151897 |
|  | MAX | 82.278481 | 92.405060 | 93.670883 | 93.670883 | 93.670883 |
|  | MIN | 82.278481 | 88.607597 | 91.139244 | 91.139244 | 91.139244 |
|  | STDEV | 0.000000 | 1.037095 | 0.719390 | 0.691987 | 0.697290 |
| 8 | AVG | 85.135132 | 90.405406 | 90.720722 | 90.810812 | 91.126127 |
|  | MAX | 85.135132 | 91.891891 | 91.891891 | 91.891891 | 93.243240 |
|  | MIN | 85.135132 | 89.189186 | 89.189186 | 90.540543 | 90.540543 |
|  | STDEV | 0.000000 | 0.740167 | 0.586719 | 0.549780 | 0.768000 |
| 9 | AVG | 86.419754 | 93.333334 | 93.744858 | 93.991772 | 94.156381 |
|  | MAX | 86.419754 | 95.061729 | 95.061729 | 95.061729 | 95.061729 |
|  | MIN | 86.419754 | 92.592590 | 92.592590 | 92.592590 | 93.827164 |
|  | STDEV | 0.000000 | 0.695363 | 0.555281 | 0.536015 | 0.555279 |
| 10 | AVG | 85.185188 | 90.370371 | 90.617285 | 90.823046 | 91.028807 |
|  | MAX | 85.185188 | 91.358025 | 91.358025 | 92.592590 | 92.592590 |
|  | MIN | 85.185188 | 88.888885 | 88.888885 | 90.123459 | 90.123459 |
|  | STDEV | 0.000000 | 0.753404 | 0.695360 | 0.701629 | 0.720113 |

Table A2. The results obtained based on the Breast-cancer dataset.

| Fold | Index | SVM | $F_{25}$ | $F_{50}$ | $F_{75}$ | $F_{100}$ |
|---|---|---|---|---|---|---|
| 1 | AVG | 95.588234 | 98.529411 | 98.529411 | 98.529411 | 98.578431 |
|   | MAX | 95.588234 | 98.529411 | 98.529411 | 98.529411 | 100.000000 |
|   | MIN | 95.588234 | 98.529411 | 98.529411 | 98.529411 | 98.529411 |
|   | STDEV | 0.000000 | 0.000000 | 0.000000 | 0.000000 | 0.268492 |
| 2 | AVG | 100.000000 | 100.000000 | 100.000000 | 100.000000 | 100.000000 |
|   | MAX | 100.000000 | 100.000000 | 100.000000 | 100.000000 | 100.000000 |
|   | MIN | 100.000000 | 100.000000 | 100.000000 | 100.000000 | 100.000000 |
|   | STDEV | 0.000000 | 0.000000 | 0.000000 | 0.000000 | 0.000000 |
| 3 | AVG | 95.588234 | 97.303921 | 97.499999 | 97.598038 | 97.745097 |
|   | MAX | 95.588234 | 98.529411 | 98.529411 | 98.529411 | 98.529411 |
|   | MIN | 95.588234 | 97.058823 | 97.058823 | 97.058823 | 97.058823 |
|   | STDEV | 0.000000 | 0.557425 | 0.685429 | 0.720783 | 0.746201 |
| 4 | AVG | 100.000000 | 100.000000 | 100.000000 | 100.000000 | 100.000000 |
|   | MAX | 100.000000 | 100.000000 | 100.000000 | 100.000000 | 100.000000 |
|   | MIN | 100.000000 | 100.000000 | 100.000000 | 100.000000 | 100.000000 |
|   | STDEV | 0.000000 | 0.000000 | 0.000000 | 0.000000 | 0.000000 |
| 5 | AVG | 100.000000 | 100.000000 | 100.000000 | 100.000000 | 100.000000 |
|   | MAX | 100.000000 | 100.000000 | 100.000000 | 100.000000 | 100.000000 |
|   | MIN | 100.000000 | 100.000000 | 100.000000 | 100.000000 | 100.000000 |
|   | STDEV | 0.000000 | 0.000000 | 0.000000 | 0.000000 | 0.000000 |
| 6 | AVG | 94.117645 | 97.058823 | 97.058823 | 97.058823 | 97.058823 |
|   | MAX | 94.117645 | 97.058823 | 97.058823 | 97.058823 | 97.058823 |
|   | MIN | 94.117645 | 97.058823 | 97.058823 | 97.058823 | 97.058823 |
|   | STDEV | 0.000000 | 0.000000 | 0.000000 | 0.000000 | 0.000000 |
| 7 | AVG | 95.588234 | 97.892156 | 98.039215 | 98.284313 | 98.333333 |
|   | MAX | 95.588234 | 100.000000 | 100.000000 | 100.000000 | 100.000000 |
|   | MIN | 95.588234 | 97.058823 | 97.058823 | 97.058823 | 97.058823 |
|   | STDEV | 0.000000 | 0.920680 | 0.891880 | 0.870726 | 0.840216 |
| 8 | AVG | 98.529411 | 98.529411 | 98.529411 | 98.529411 | 98.529411 |
|   | MAX | 98.529411 | 98.529411 | 98.529411 | 98.529411 | 98.529411 |
|   | MIN | 98.529411 | 98.529411 | 98.529411 | 98.529411 | 98.529411 |
|   | STDEV | 0.000000 | 0.000000 | 0.000000 | 0.000000 | 0.000000 |
| 9 | AVG | 94.117645 | 98.529411 | 98.529411 | 98.529411 | 98.529411 |
|   | MAX | 94.117645 | 98.529411 | 98.529411 | 98.529411 | 98.529411 |
|   | MIN | 94.117645 | 98.529411 | 98.529411 | 98.529411 | 98.529411 |
|   | STDEV | 0.000000 | 0.000000 | 0.000000 | 0.000000 | 0.000000 |
| 10 | AVG | 97.183098 | 97.183098 | 97.183098 | 97.183098 | 97.230046 |
|   | MAX | 97.183098 | 97.183098 | 97.183098 | 97.183098 | 98.591553 |
|   | MIN | 97.183098 | 97.183098 | 97.183098 | 97.183098 | 97.183098 |
|   | STDEV | 0.000000 | 0.000000 | 0.000000 | 0.000000 | 0.257148 |

**Table A3.** The results obtained based on the Diabetes dataset.

| Fold | Index | SVM | $F_{25}$ | $F_{50}$ | $F_{75}$ | $F_{100}$ |
|---|---|---|---|---|---|---|
| 1 | AVG | 76.712326 | 83.515981 | 83.926940 | 84.018263 | 84.018263 |
|   | MAX | 76.712326 | 84.931503 | 84.931503 | 84.931503 | 84.931503 |
|   | MIN | 76.712326 | 80.821915 | 80.821915 | 80.821915 | 80.821915 |
|   | STDEV | 0.000000 | 1.271038 | 1.133810 | 1.156413 | 1.156413 |
| 2 | AVG | 76.811592 | 82.512076 | 82.946859 | 83.140096 | 83.236714 |
|   | MAX | 76.811592 | 85.507248 | 85.507248 | 85.507248 | 85.507248 |
|   | MIN | 76.811592 | 79.710144 | 81.159416 | 81.159416 | 81.159416 |
|   | STDEV | 0.000000 | 1.314764 | 1.054974 | 1.041153 | 0.983928 |
| 3 | AVG | 82.022469 | 86.404494 | 86.666666 | 86.891385 | 86.891385 |
|   | MAX | 82.022469 | 87.640450 | 87.640450 | 87.640450 | 87.640450 |
|   | MIN | 82.022469 | 85.393257 | 85.393257 | 85.393257 | 85.393257 |
|   | STDEV | 0.000000 | 0.743555 | 0.765669 | 0.681437 | 0.681437 |
| 4 | AVG | 76.470589 | 79.558825 | 79.705884 | 79.803923 | 79.852943 |
|   | MAX | 76.470589 | 80.882355 | 80.882355 | 80.882355 | 80.882355 |
|   | MIN | 76.470589 | 79.411766 | 79.411766 | 79.411766 | 79.411766 |
|   | STDEV | 0.000000 | 0.448719 | 0.598292 | 0.661436 | 0.685429 |
| 5 | AVG | 71.232880 | 71.278542 | 71.324203 | 71.369865 | 71.369865 |
|   | MAX | 71.232880 | 72.602737 | 72.602737 | 72.602737 | 72.602737 |
|   | MIN | 71.232880 | 71.232880 | 71.232880 | 71.232880 | 71.232880 |
|   | STDEV | 0.000000 | 0.250101 | 0.347544 | 0.417983 | 0.417983 |
| 6 | AVG | 68.292686 | 71.138209 | 71.910567 | 72.154470 | 72.154470 |
|   | MAX | 68.292686 | 73.170731 | 73.170731 | 73.170731 | 73.170731 |
|   | MIN | 68.292686 | 69.512192 | 70.731705 | 70.731705 | 70.731705 |
|   | STDEV | 0.000000 | 0.924510 | 0.815458 | 0.852356 | 0.852356 |
| 7 | AVG | 82.558144 | 82.558144 | 82.596903 | 82.596903 | 82.635663 |
|   | MAX | 82.558144 | 82.558144 | 83.720932 | 83.720932 | 83.720932 |
|   | MIN | 82.558144 | 82.558144 | 82.558144 | 7.000000 | 82.558144 |
|   | STDEV | 0.000000 | 0.000000 | 0.212295 | 0.212295 | 0.295009 |
| 8 | AVG | 71.428574 | 74.025972 | 74.329003 | 74.632034 | 74.805194 |
|   | MAX | 71.428574 | 75.324677 | 75.324677 | 75.324677 | 75.324677 |
|   | MIN | 71.428574 | 71.428574 | 72.727272 | 72.727272 | 72.727272 |
|   | STDEV | 0.000000 | 1.023167 | 0.881703 | 0.742010 | 0.731485 |
| 9 | AVG | 80.519478 | 83.246752 | 83.679652 | 83.852811 | 83.939391 |
|   | MAX | 80.519478 | 84.415581 | 84.415581 | 84.415581 | 84.415581 |
|   | MIN | 80.519478 | 81.818184 | 81.818184 | 81.818184 | 83.116882 |
|   | STDEV | 0.000000 | 0.859431 | 0.738077 | 0.738077 | 0.636534 |
| 10 | AVG | 78.378380 | 84.009009 | 84.594594 | 84.909910 | 85.000000 |
|   | MAX | 78.378380 | 87.837837 | 87.837837 | 87.837837 | 87.837837 |
|   | MIN | 78.378380 | 82.432434 | 82.432434 | 82.432434 | 82.432434 |
|   | STDEV | 0.000000 | 1.378265 | 1.610712 | 1.590385 | 1.677734 |

**Table A4.** The results obtained based on the Fourclass dataset.

| Fold | Index | SVM | $F_{25}$ | $F_{50}$ | $F_{75}$ | $F_{100}$ |
|---|---|---|---|---|---|---|
| 1 | AVG | 80.232559 | 86.240310 | 87.015504 | 87.248062 | 87.403101 |
|   | MAX | 80.232559 | 88.372093 | 88.372093 | 88.372093 | 88.372093 |
|   | MIN | 80.232559 | 83.720932 | 83.720932 | 84.883720 | 84.883720 |
|   | STDEV | 0.000000 | 1.674868 | 1.262130 | 1.201573 | 1.104529 |
| 2 | AVG | 79.000000 | 81.133333 | 81.333333 | 81.433333 | 81.533333 |
|   | MAX | 79.000000 | 82.000000 | 82.000000 | 82.000000 | 82.000000 |
|   | MIN | 79.000000 | 81.000000 | 81.000000 | 81.000000 | 81.000000 |
|   | STDEV | 0.000000 | 0.345746 | 0.479463 | 0.504007 | 0.507416 |
| 3 | AVG | 75.862068 | 81.072795 | 81.417622 | 81.494250 | 81.570878 |
|   | MAX | 75.862068 | 82.758621 | 82.758621 | 82.758621 | 82.758621 |
|   | MIN | 75.862068 | 78.160919 | 78.160919 | 78.160919 | 80.459770 |
|   | STDEV | 0.000000 | 0.839420 | 0.744504 | 0.698190 | 0.367634 |
| 4 | AVG | 81.818184 | 83.863637 | 84.166668 | 84.356062 | 84.659091 |
|   | MAX | 81.818184 | 86.363632 | 86.363632 | 86.363632 | 86.363632 |
|   | MIN | 81.818184 | 82.954544 | 82.954544 | 82.954544 | 82.954544 |
|   | STDEV | 0.000000 | 0.811801 | 0.891949 | 0.879379 | 0.882748 |
| 5 | AVG | 82.857140 | 84.285713 | 84.285713 | 84.285713 | 84.333332 |
|   | MAX | 82.857140 | 84.285713 | 84.285713 | 84.285713 | 85.714287 |
|   | MIN | 82.857140 | 84.285713 | 84.285713 | 84.285713 | 84.285713 |
|   | STDEV | 0.000000 | 0.000000 | 0.000000 | 0.000000 | 0.260821 |
| 6 | AVG | 81.914894 | 82.978722 | 83.014183 | 83.049644 | 83.085105 |
|   | MAX | 81.914894 | 82.978722 | 84.042557 | 84.042557 | 84.042557 |
|   | MIN | 81.914894 | 82.978722 | 82.978722 | 82.978722 | 82.978722 |
|   | STDEV | 0.000000 | 0.000000 | 0.194229 | 0.269904 | 0.324607 |
| 7 | AVG | 85.897438 | 85.897438 | 85.897438 | 85.897438 | 85.897438 |
|   | MAX | 85.897438 | 85.897438 | 85.897438 | 85.897438 | 85.897438 |
|   | MIN | 85.897438 | 85.897438 | 85.897438 | 85.897438 | 85.897438 |
|   | STDEV | 0.000000 | 0.000000 | 0.000000 | 0.000000 | 0.000000 |
| 8 | AVG | 71.428574 | 77.752811 | 77.940077 | 74.632034 | 78.127343 |
|   | MAX | 71.428574 | 78.651688 | 78.651688 | 75.324677 | 78.651688 |
|   | MIN | 71.428574 | 77.528091 | 77.528091 | 72.727272 | 77.528091 |
|   | STDEV | 0.000000 | 0.457122 | 0.550711 | 0.742010 | 0.570131 |
| 9 | AVG | 80.519478 | 84.573643 | 85.155038 | 83.852811 | 85.465115 |
|   | MAX | 80.519478 | 86.046509 | 86.046509 | 84.415581 | 86.046509 |
|   | MIN | 80.519478 | 81.395348 | 82.558144 | 81.818184 | 83.720932 |
|   | STDEV | 0.000000 | 1.363349 | 1.043759 | 0.738077 | 0.732235 |
| 10 | AVG | 78.378380 | 89.285713 | 89.285713 | 84.909910 | 89.285713 |
|    | MAX | 78.378380 | 89.285713 | 89.285713 | 87.837837 | 89.285713 |
|    | MIN | 78.378380 | 89.285713 | 89.285713 | 82.432434 | 89.285713 |
|    | STDEV | 0.000000 | 0.000000 | 0.000000 | 1.590385 | 0.000000 |

Table A5. The results obtained based on the Heart Disease dataset.

| Fold | Index | SVM | $F_{25}$ | $F_{50}$ | $F_{75}$ | $F_{100}$ |
|---|---|---|---|---|---|---|
| 1 | AVG | 80.000000 | 88.400000 | 88.533333 | 88.933333 | 89.066667 |
|   | MAX | 80.000000 | 92.000000 | 92.000000 | 92.000000 | 92.000000 |
|   | MIN | 80.000000 | 88.000000 | 88.000000 | 88.000000 | 88.000000 |
|   | STDEV | 0.000000 | 1.220514 | 1.382984 | 1.720732 | 1.799106 |
| 2 | AVG | 82.758621 | 96.321842 | 96.436785 | 96.551727 | 96.551727 |
|   | MAX | 82.758621 | 96.551727 | 96.551727 | 96.551727 | 96.551727 |
|   | MIN | 82.758621 | 93.103447 | 93.103447 | 96.551727 | 96.551727 |
|   | STDEV | 0.000000 | 0.874857 | 0.629567 | 0.000000 | 0.000000 |
| 3 | AVG | 82.758621 | 89.655174 | 89.770117 | 89.770117 | 89.770117 |
|   | MAX | 82.758621 | 89.655174 | 93.103447 | 93.103447 | 93.103447 |
|   | MIN | 82.758621 | 89.655174 | 89.655174 | 89.655174 | 89.655174 |
|   | STDEV | 0.000000 | 0.000000 | 0.629566 | 0.629566 | 0.629566 |
| 4 | AVG | 88.235291 | 92.745096 | 93.235292 | 93.431371 | 93.431371 |
|   | MAX | 88.235291 | 94.117645 | 94.117645 | 94.117645 | 94.117645 |
|   | MIN | 88.235291 | 91.176468 | 91.176468 | 91.176468 | 91.176468 |
|   | STDEV | 0.000000 | 1.492401 | 1.370858 | 1.265245 | 1.265245 |
| 5 | AVG | 100.00000 | 100.000000 | 100.000000 | 100.000000 | 100.000000 |
|   | MAX | 100.00000 | 100.000000 | 100.000000 | 100.000000 | 100.000000 |
|   | MIN | 100.00000 | 100.000000 | 100.000000 | 100.000000 | 100.000000 |
|   | STDEV | 0.000000 | 0.000000 | 0.000000 | 0.000000 | 0.000000 |
| 6 | AVG | 80.000000 | 93.500000 | 94.666667 | 94.833333 | 95.000000 |
|   | MAX | 80.000000 | 95.000000 | 95.000000 | 95.000000 | 95.000000 |
|   | MIN | 80.000000 | 90.000000 | 90.000000 | 90.000000 | 95.000000 |
|   | STDEV | 0.000000 | 2.330458 | 1.268541 | 0.912871 | 0.000000 |
| 7 | AVG | 82.608696 | 95.652176 | 95.797103 | 95.797103 | 95.797103 |
|   | MAX | 82.608696 | 100.000000 | 100.000000 | 100.000000 | 100.000000 |
|   | MIN | 82.608696 | 91.304352 | 95.652176 | 95.652176 | 95.652176 |
|   | STDEV | 0.000000 | 1.141795 | 0.793800 | 0.793800 | 0.793800 |
| 8 | AVG | 89.285713 | 89.880951 | 90.476189 | 90.714284 | 91.071426 |
|   | MAX | 89.285713 | 92.857140 | 92.857140 | 92.857140 | 92.857140 |
|   | MIN | 89.285713 | 89.285713 | 89.285713 | 89.285713 | 89.285713 |
|   | STDEV | 0.000000 | 1.353746 | 1.712368 | 1.779545 | 1.816240 |
| 9 | AVG | 75.000000 | 86.203707 | 86.388892 | 86.481484 | 86.481484 |
|   | MAX | 75.000000 | 88.888885 | 88.888885 | 88.888885 | 88.888885 |
|   | MIN | 75.000000 | 86.111115 | 86.111115 | 86.111115 | 86.111115 |
|   | STDEV | 0.000000 | 0.507149 | 0.847577 | 0.960403 | 0.960403 |
| 10 | AVG | 84.615387 | 88.717950 | 88.974361 | 89.230771 | 89.615386 |
|    | MAX | 84.615387 | 92.307693 | 92.307693 | 92.307693 | 92.307693 |
|    | MIN | 84.615387 | 88.461540 | 88.461540 | 88.461540 | 88.461540 |
|    | STDEV | 0.000000 | 0.975800 | 1.329792 | 1.564762 | 1.792660 |